\documentclass[runningheads]{llncs}

\usepackage{eccv}

\usepackage{eccvabbrv}

\usepackage{graphicx}
\usepackage{booktabs}

\usepackage[accsupp]{axessibility}  

\usepackage{hyperref}

\usepackage{orcidlink}

\usepackage{wrapfig}
\usepackage{algorithm}
\usepackage{algpseudocode}

\begin{document}

\title{AR-CoPO: Align Autoregressive Video Generation  with Contrastive Policy Optimization} 

\titlerunning{AR-CoPO}

\author{
\bf Dailan He$^{1,2}$~~~ Guanlin Feng$^{2}$~~~~ Xingtong Ge$^{2,3}$~~~~  Yi Zhang$^{2}$~~~~ \\
\bf Bingqi Ma$^{2}$~~~~ Guanglu Song$^{2}$~~~~ Yu Liu$^{2}$~~~~ Hongsheng Li$^{1,4}$
}

\authorrunning{He.~et al.}

\institute{CUHK MMLab, China  \email{hedailan@link.cuhk.edu.hk}
\and
Vivix Group Limited, China \and HKUST, China \and Shenzhen Loop Area Institute, China
}

\maketitle

\begin{abstract}
  Streaming autoregressive (AR) video generators combined with few-step distillation achieve low-latency, high-quality synthesis, yet remain difficult to align via reinforcement learning from human feedback (RLHF). Existing SDE-based GRPO methods face challenges in this setting: few-step ODEs and consistency model samplers deviate from standard flow-matching ODEs, and their short, low-stochasticity trajectories are highly sensitive to initialization noise, rendering intermediate SDE exploration ineffective. We propose \textbf{AR-CoPO} (\textbf{A}uto\textbf{R}egressive \textbf{Co}ntrastive \textbf{P}olicy \textbf{O}ptimization), a framework that adapts the Neighbor GRPO contrastive perspective to streaming AR generation. AR-CoPO introduces chunk-level alignment via a forking mechanism that constructs neighborhood candidates at a randomly selected chunk, assigns sequence-level rewards, and performs localized GRPO updates. We further propose a semi-on-policy training strategy that complements on-policy exploration with exploitation over a replay buffer of reference rollouts, improving generation quality across domains. Experiments on Self-Forcing demonstrate that AR-CoPO improves both out-of-domain generalization and in-domain human preference alignment over the baseline, providing evidence of genuine alignment rather than reward hacking.
  \keywords{Autoregressive Video Generation \and RLHF}
  \end{abstract}
  
  \begin{figure}
    \centering
    \includegraphics[width=0.86\linewidth]{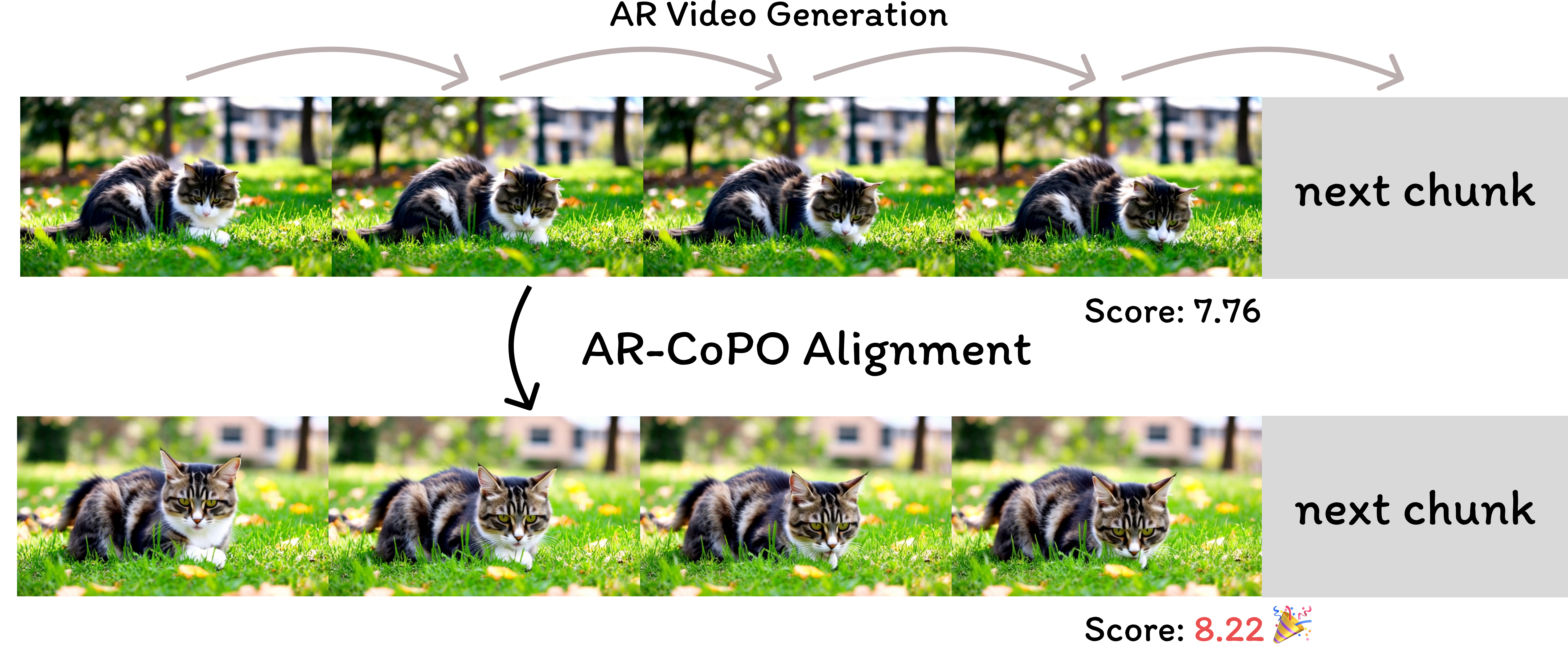}
    \caption{AR-CoPO is a reinforcement learning for human preference (RLHF) method, aligning few-step autoregressive (AR) video generative models to better sample quality.}
    \label{fig:teaser}
  \end{figure}
  
  \section{Introduction}
  \label{sec:intro}

  Diffusion and flow-matching models~\cite{ho2020ddpm, rombach2022ldm, lipman2022flowmatching} have achieved remarkable progress in image and video synthesis~\cite{labs2025flux1kontextflowmatching,rombach2022ldm,kong2024hunyuanvideo,wu2025hunyuanvideo15,zheng2024opensora,zhang2025waver}, delivering high-quality and diverse results. Despite their success, the inference cost of bidirectional generation typically scales linearly with both the number of sampling steps and the target video length. This scaling behavior makes it difficult to deploy these models in low-latency, variable-length, and streaming-generation scenarios. To address this, a growing line of work~\cite{yin2025causvid,huang2025selfforcing,zhu2026causalforcing,ge2026salt,yang2025longlive} distills powerful bidirectional video models into causal generators that operate in an autoregressive (AR) and chunk-wise manner. In addition, techniques like distribution matching distillation (DMD)~\cite{yin2024dmd,yin2024dmd2} compress the sampling process into just a few steps. These are often implemented via few-step ODE solvers~\cite{song2020ddim, lu2022dpm} or consistency models, while engineering optimizations such as KV caching further improve throughput and responsiveness.
  
  However, the combination of a streaming AR structure and few-step distillation introduces significant challenges for post-training alignment, particularly for reinforcement learning from human feedback (RLHF)~\cite{christiano2017rlhf, stiennon2020rlhf, ouyang2022rlhfinstruct}. RLHF is essential for controllable, high-quality video generation. Similar to language modeling, video generators are often aligned to reward models or verifiers that capture instruction following, subject consistency, motion plausibility, and aesthetics. For flow-matching style generators, policy-gradient post-training (\textit{e.g.}, GRPO-like objectives~\cite{shao2024deepseekmath, schulman2017ppo}) provides a natural route by viewing the sampling process as policy rollouts and optimizing the induced distribution using reward feedback. 
  
  \begin{figure}[t]
    \centering
    \begin{subfigure}{0.33\linewidth}
      \includegraphics[width=0.99\textwidth]{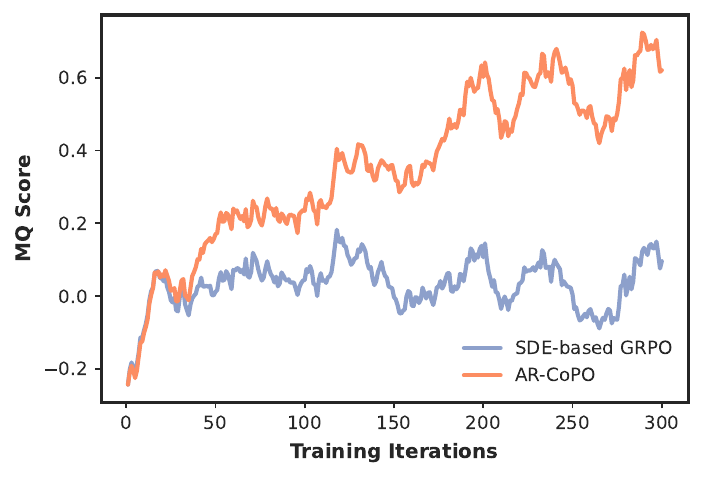}
        \caption*{}
        \label{fig:sde-vs-arcopo:curve}
    \end{subfigure}
    \begin{subfigure}{0.6\linewidth}
      \includegraphics[width=0.99\textwidth]{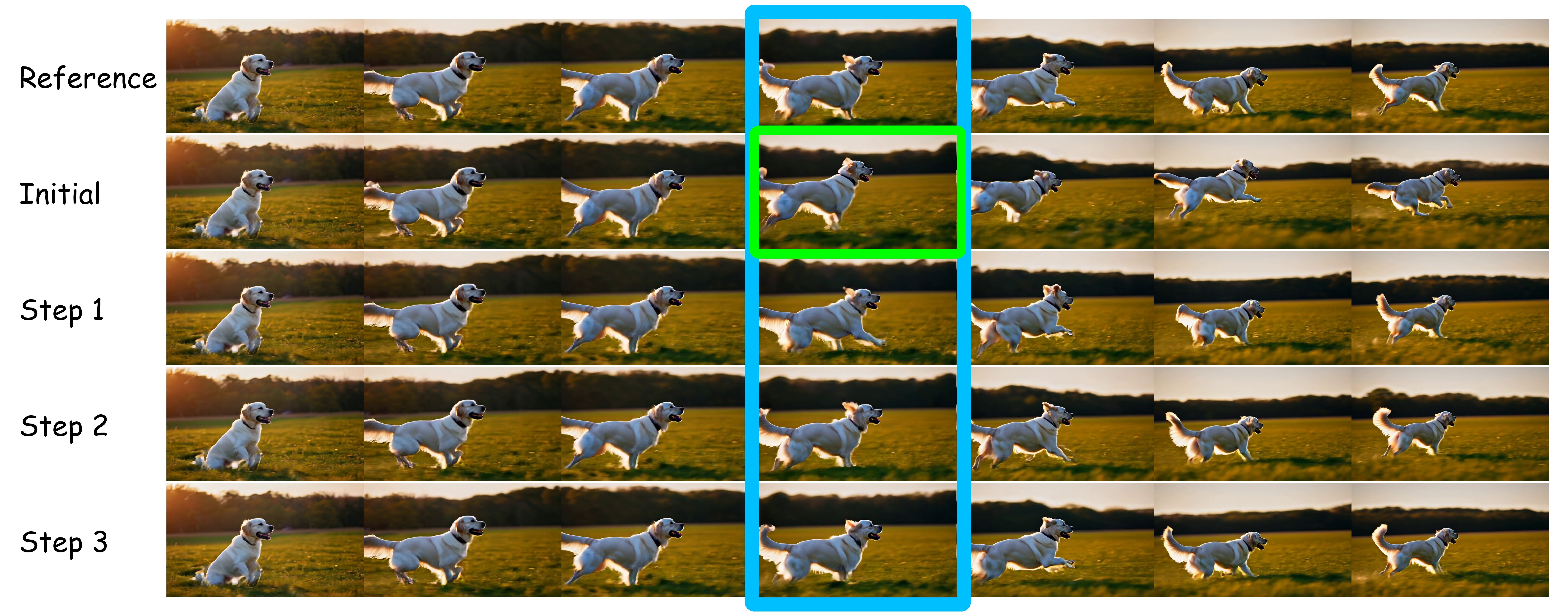}
        \caption*{}
        \label{fig:sde-vs-arcopo:randomness}
    \end{subfigure}
    \caption{\textbf{Left:} Training curves comparing SDE-based GRPO and AR-CoPO on Self-Forcing. SDE-based GRPO fails to improve the reward, while AR-CoPO consistently achieves higher scores throughout training. \textbf{Right:} Perturbing only the intermediate CM solver noise (Rows 3–5) produces nearly identical outputs, whereas replacing the initial noise (Row 2) causes significant variation, confirming that few-step AR models (\eg Self-Forcing~\cite{huang2025selfforcing}) are near-deterministic and driven primarily by initial noise.}
    \label{fig:sde-vs-arcopo}
  \end{figure}

  Many practical GRPO variants designed for flow-matching models, however, hinge on an important implementation choice: converting deterministic ODE sampling into a stochastic SDE form to introduce a Markov decision process (MDP). Applying these SDE-based GRPO methods~\cite{xue2025dancegrpo, liu2025flowgrpo, li2025mixgrpo, li2025branchgrpo, wang2025prefgrpo} to few-step streaming generators presents significant challenges. First, few-step generators—often formulated as distilled few-step ODEs or consistency models—deviate from standard flow-matching ODEs, making them inherently difficult to train using SDE-based methods designed for continuous flow matching. Second, because these models follow short sampling trajectories with limited stochasticity, they are highly sensitive to initialization noise and model approximation errors. In contrast, SDE-based approaches heavily rely on intermediate noise injections to guide exploration while typically keeping the initial noise frozen, leading to a fundamental mismatch. This paper, therefore, addresses a critical question: How can we effectively align streaming AR video generators using GRPO-like objectives?
  
  We draw inspiration from the recently proposed Neighbor GRPO~\cite{he2025neighborgrpo} perspective, which offers a novel reinterpretation: the SDE-GRPO update can be mathematically reformulated as a distance-driven contrastive objective over a set of neighbor candidate trajectories. Crucially, this view suggests that one can obtain informative preference signals without relying on stochastic SDE exploration during sampling. Instead, exploration can be entirely controlled during training by constructing neighborhood candidates around the initial noise of an ODE sampler and defining a softmax distance-based surrogate transition distribution over these candidates.
  
  Based on this insight, we propose an \textbf{A}uto\textbf{R}egressive \textbf{Co}ntrastive \textbf{P}olicy \textbf{O}ptimization (\textbf{AR-CoPO}) framework for the post-training alignment of streaming AR video generators. To match the structural properties of streaming AR generation, we introduce \textbf{chunk-level alignment} as the fundamental unit of neighborhood construction and optimization. This establishes a chunk-level action space that responds to sequence-level rewards, enabling natural and localized credit assignment. Furthermore, in addition to standard on-policy training that encourages exploration, we adopt a \textbf{semi-on-policy} training strategy under the AR-CoPO objective to enhance exploitation. By combining these two training paradigms, we effectively improve both out-of-domain performance and overall generation quality. In summary, our contributions are:
  \begin{enumerate}

  \item We propose the AR-CoPO framework, which enables the RLHF of streaming video generators using a contrastive objective over neighborhood candidates.
  \item We introduce a chunk-level action space that allows for natural credit assignment from sequence-level rewards, producing localized and controllable reward signals.
  \item We propose a semi-on-policy training strategy under the AR-CoPO objective to enhance exploitation. Combined with standard on-policy exploration, this effectively improves both out-of-domain performance and overall generation quality.
  \end{enumerate}

  \section{Related Works}
  
  \subsection{Autoregressive Video Generation}
  
  Flow matching (FM)~\cite{lipman2022flowmatching} and related probability flow ODEs provide a deterministic alternative to stochastic diffusion sampling, enabling fast generation via few-step ODE solvers when combined with distillation techniques like distribution matching distillation (DMD)~\cite{yin2024dmd,yin2024dmd2}. While ODE determinism enables efficient and stable inference, it limits the stochastic exploration typically required for RL-based alignment. Our work targets alignment mechanisms that enable controllable exploration during training while preserving fast deterministic inference.
  
  To support streaming applications and reduce latency, recent methods distill bidirectional video generators into \emph{causal} autoregressive (AR) generators that synthesize videos chunk by chunk~\cite{yin2025causvid,huang2025selfforcing,zhu2026causalforcing,yang2025longlive}. However, the AR setting introduces new challenges, such as cross-chunk error accumulation and exposure bias due to the mismatch between training and inference contexts. These challenges motivate our focus on developing stable and localized alignment methods tailored for streaming AR generation and short-horizon few-step ODE inference.
  
  \subsection{Post-training Flow Alignment and Neighbor GRPO}
  
  Post-training alignment for generative models typically leverages human preferences or computable rewards to improve instruction following, aesthetics, and temporal consistency. For diffusion and flow-based generators, reinforcement-learning style objectives can be applied by viewing sampling as a policy rollout and optimizing the induced distribution using policy gradients, often with variants that avoid training an explicit critic (e.g., GRPO-like methods~\cite{shao2024deepseekmath}). A common practical choice is to convert deterministic ODE sampling into a stochastic SDE process to inject exploration and to obtain lower-variance updates~\cite{xue2025dancegrpo, liu2025flowgrpo, li2025mixgrpo, li2025branchgrpo, wang2025prefgrpo}, but this conversion can restrict solver choices and tends to be more effective with longer trajectories. Alternative approaches align diffusion models via direct preference optimization or reward-supervised fine-tuning~\cite{wallace2024diffusiondpo, wu2024drtune}.
  
  Neighbor GRPO~\cite{he2025neighborgrpo} provides an alternative perspective by reinterpreting SDE-based GRPO updates as a distance-driven contrastive objective over a neighborhood of candidate trajectories. By constructing neighbors via perturbations of the initial noise and defining a training-time surrogate transition distribution based on softmax distances in trajectory/latent space, it enables informative preference signals while keeping inference-time sampling purely ODE and compatible with high-order solvers. Our method follows this high-level principle---\emph{training-time controllable exploration with inference-time determinism}---but adapts it to the streaming AR video setting. In particular, we formulate alignment at the chunk level to better match autoregressive structure, aiming to provide more localized and stable signals under cross-chunk error propagation, and we introduce a semi-on-policy strategy to improve data reuse and wall-clock efficiency.
  
  \section{Methodology}
  \subsection{Preliminary: Neighbor GRPO}
  \label{sec:preliminary}
  
  Neighbor GRPO~\cite{he2025neighborgrpo} aligns flow-based generative models by reinterpreting SDE-based policy optimization as a distance-driven contrastive learning objective, and adopts a contrastive policy optimization (CoPO). To preserve the fast, deterministic inference of ODE sampling, it avoids stochastic SDE conversions. Instead, it constructs a neighborhood of candidate trajectories by perturbing a shared initial noise $\epsilon^* \sim \mathcal{N}(0, I)$:
  \begin{equation}
  \label{eq:neighbor_noise}
  \epsilon^{(i)} = \sqrt{1-\sigma^2}\,\epsilon^* + \sigma\,\delta^{(i)}, \quad \delta^{(i)} \sim \mathcal{N}(0, I), \quad i=1,\ldots,G,
  \end{equation}
  where $\sigma \in (0, 1)$ controls the exploration radius. These noises are deterministically rolled out using a reference policy via an ODE solver, and intermediate latents $\{x_t^{(i)}\}_{i=1}^G$ at a chosen timestep $t$ are collected as candidates. 
  
  To enable policy gradient updates, Neighbor GRPO defines a surrogate training-time transition distribution based on the distance between an anchor latent $x_t^{(\theta)}$ (produced by the active policy $\theta$ from the same timestep) and the candidates:
  \begin{equation}
  \label{eq:neighbor_surrogate}
  d^{(i)}=\left\|x_t^{(i)}-x_t^{(\theta)}\right\|_2^2,\qquad
  \pi_\theta(i) = \frac{\exp\!\left(-d^{(i)}/\tau \right)}{\sum_{k=1}^G \exp\!\left(-d^{(k)}/\tau \right)},
  \end{equation}
  where $\tau$ is a temperature hyperparameter. Given rewards $\{r^{(i)}\}_{i=1}^G$ for the candidates, it computes group-normalized advantages $A^{(i)} = \frac {r^{(i)} - \bar{r}} {\sigma_r}$.
  Plugging the surrogate policy and advantages into the GRPO objective, the model is optimized to maximize:
  \begin{equation}
  \label{eq:neighbor_obj}
  J(\theta) = \frac{1}{G}\sum_{i=1}^G \min\left( \frac{\pi_\theta(i)}{\pi_{\text{old}}(i)} A^{(i)}, \text{clip}\left(\frac{\pi_\theta(i)}{\pi_{\text{old}}(i)}, 1-\epsilon, 1+\epsilon\right) A^{(i)} \right).
  \end{equation}
  This objective naturally pulls the anchor towards candidates with positive advantages and pushes it away from those with negative advantages. Furthermore, Neighbor GRPO employs symmetric anchor sampling to reduce training costs by treating multiple candidates as anchors without extra forward passes.
  
  \subsection{AR-CoPO}
  
  \begin{figure}[t]
    \centering
    \includegraphics[width=0.98\linewidth]{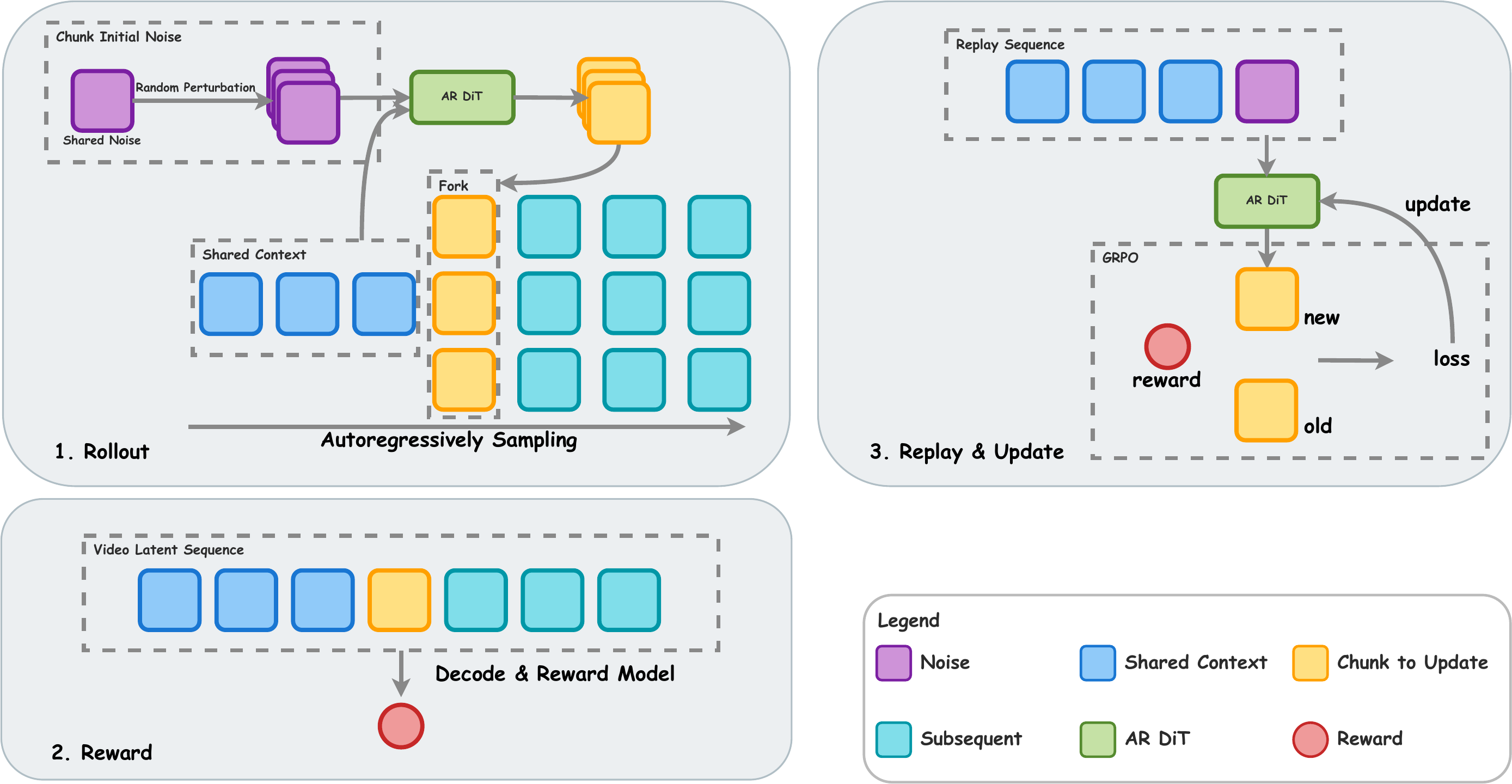}
    \caption{The AR-CoPO training pipeline. \textbf{(1) Rollout:} The model autoregressively generates a shared context up to a randomly selected pivot chunk $p$. At chunk $p$, the base initial noise is perturbed into $G$ neighbors; each neighbor is forked into an independent branch and autoregressively completed to produce a full video sequence. \textbf{(2) Reward:} Each completed sequence is decoded and scored by a reward model, yielding a sequence-level reward per branch. \textbf{(3) Replay \& Update:} The saved pivot-chunk trajectories are replayed through the current policy; distances between current and old $\hat{x}_0$ predictions induce surrogate policy ratios, which are used in a clipped GRPO update confined to the pivot chunk.}
    \label{fig:arcopo-framework}
  \end{figure}
  
  In this section we introduce AR-CoPO. As illustrated in Figure~\ref{fig:arcopo-framework}, the training pipeline consists of three phases: (1) rolling out neighborhood candidates by forking at a randomly selected pivot chunk, (2) computing sequence-level rewards for each completed branch, and (3) replaying the pivot chunk through the current policy and performing a contrastive GRPO update. We detail each component below.
  
  \subsubsection{Chunk-level Alignment via Forking}
  Due to the streaming nature of autoregressive (AR) generation, naive application of sequence-level GRPO is prohibitively expensive and suffers from severe credit assignment issues. To address this, we propose a chunk-level alignment strategy that performs action space sampling (forking) at a randomly selected chunk, and evaluates the generation via a sequence-level reward.
  
  Specifically, for a target sequence of length $L$ chunks, each optimization iteration proceeds as follows:
  \begin{enumerate}
    \item \textbf{Shared Context Generation}: We randomly sample a pivot chunk index $p \in \{1, \dots, L\}$. The model sequentially generates the first $p-1$ chunks to establish a shared historical context $h_{p-1}$ (e.g., cached KV states).
    \item \textbf{Action Space Forking}: At the $p$-th chunk, we branch the generation to perform action space sampling. Based on a shared base initial noise $\epsilon_p^*$, we construct a group of $G$ perturbed neighboring noises $\{\epsilon_p^{(i)}\}_{i=1}^G$ following Eq.~\ref{eq:neighbor_noise}. For each branch $i$, the model completes the $T$-step denoising generation to produce the chunk latent $x_p^{(i)}$. The states along this $T$-step trajectory are stored in a replay buffer.
    \item \textbf{Rollout and Sequence-Level Reward}: For each of the $G$ branches, the model deterministically generates the remaining $L-p$ chunks (without further perturbation). After the full sequence is completed, we compute a \emph{sequence-level} reward $r^{(i)}$ for each branch.
  \end{enumerate}

  \textbf{Controlled noise sharing.}
  Crucially, within each training iteration, the only source of randomness that \emph{differs} across branches is the initial noise of the pivot chunk $\epsilon_p^{(i)}$.
  All other noise sequences are shared identically across branches: the initial noises of every non-pivot chunk and all CM solver noises at every denoising timestep within every chunk are drawn once and reused across all $G$ branches.
  This design ensures that the $G$ completed sequences are identical everywhere except in the content generated at chunk $p$, so any reward difference $r^{(i)} - r^{(j)}$ is cleanly attributable to the choice of $\epsilon_p^{(i)}$ and the resulting latent $x_p^{(i)}$, with no confounding stochasticity from later generation stages.
  
  During the policy update phase, we retrieve the saved trajectories of the $p$-th chunk from the replay buffer. We compute the advantages $A^{(i)}$ using the sequence-level rewards $r^{(i)}$. Then, utilizing the distance-induced surrogate policy $\pi_\theta(i\mid s_p)$ (where distances are computed using the chunk latents $x_p$), we perform a standard Neighbor GRPO parameter update optimizing Eq.~\ref{eq:neighbor_obj} restricted to the $p$-th chunk.
  
  This chunk-level forking mechanism offers two major advantages. First, it significantly reduces training costs: during replay-style gradient updates, backpropagation is strictly confined to the $T$ steps of the single $p$-th chunk, bypassing the prohibitive cost of full-sequence multi-branch backpropagation. Second, it provides a stable and localized credit assignment signal: by isolating all inter-branch randomness to the initial noise of chunk $p$ while sharing all other noise sources across the group (see above), reward differences among branches are unconfounded and can be directly attributed to the generation choices made at that specific chunk.

  \subsubsection{CoPO for Consistency Model Alignment}

  The forking mechanism described above provides a general chunk-level alignment framework, but leaves open the question of how to define the surrogate transition distribution $\pi_\theta(i \mid s_p)$ for different sampler types.
  Neighbor GRPO was originally designed for flow-matching (FM) models: it measures distances between candidate latents $x_t^{(i)}$ produced by a deterministic ODE solver at an intermediate timestep $t$ (Eq.~\ref{eq:neighbor_surrogate}), which is a natural choice when the model is a continuous-time velocity field.
  For FM-based AR generators, this original distance definition applies directly within our chunk-level framework.

  For consistency models (CMs) such as Self-Forcing~\cite{huang2025selfforcing}, however, the ODE-solver distance is not well suited.
  CMs do not follow a standard DDIM/ODE trajectory; their key operation is a one-step mapping from a noisy latent directly to a clean prediction $\hat{x}_0$.
  Measuring distances in the intermediate $x_t$ space conflates noise scale with semantic content and is therefore less informative.
  We instead apply the same CoPO principle but define the distance in $\hat{x}_0$ prediction space using the CM one-step prediction $\hat x_{0,t}=F_\theta(x_t,h_{t-1},t)$:
  \begin{equation}
  d_{0,t}^{(i)}=\left\|\hat x_{0,t}^{(i)}-\hat x_{0,t}^{(\theta)}\right\|_2^2,\qquad
  \pi_\theta(i\mid s_t)=\frac{\exp\!\left(-d_{0,t}^{(i)}/\tau_0\right)}{\sum_{k=1}^{G}\exp\!\left(-d_{0,t}^{(k)}/\tau_0\right)},
  \end{equation}
  where $\hat x_{0,t}^{(i)}=F_{\theta_{\text{old}}}(x_t^{(i)},h_{t-1},t)$ is produced by the old parameters on candidate inputs, $\hat x_{0,t}^{(\theta)}$ is the current prediction on the anchor, and $\tau_0$ is the temperature.

\begin{algorithm}[t]
\caption{AR-CoPO Training (one iteration)}
\label{alg:arcopo}
\begin{algorithmic}[1]
\Require Policy $\theta$, reward $r(\cdot)$, sequence length $L$, group size $G$
\State Sample pivot $p \sim \mathrm{Uniform}(1,L)$
\State Generate shared context $h_{p-1}$ by running $\theta$ on chunks $1,\ldots,p{-}1$
\For{$i = 1,\ldots,G$} \Comment{\textit{Fork at chunk $p$}}
  \State $\epsilon_p^{(i)} \leftarrow \sqrt{1-\sigma^2}\,\epsilon_p^* + \sigma\,\delta^{(i)},\quad \delta^{(i)}\sim\mathcal{N}(0,I)$
  \State Denoise chunk $p$ from $\epsilon_p^{(i)}$; complete remaining chunks; compute $r^{(i)}$
\EndFor
\State $A^{(i)} \leftarrow (r^{(i)} - \bar{r})/\sigma_r$
\State Replay chunk $p$: compute $\pi_\theta(i) \propto \exp(-\|\hat{x}_{0}^{(i)}-\hat{x}_{0}^{(\theta)}\|^2/\tau_0)$
\State Update $\theta$ via GRPO (Eq.~\ref{eq:neighbor_obj}) on chunk $p$ only
\end{algorithmic}
\end{algorithm}

\begin{figure}[t]
    \centering
    \includegraphics[width=0.7\linewidth]{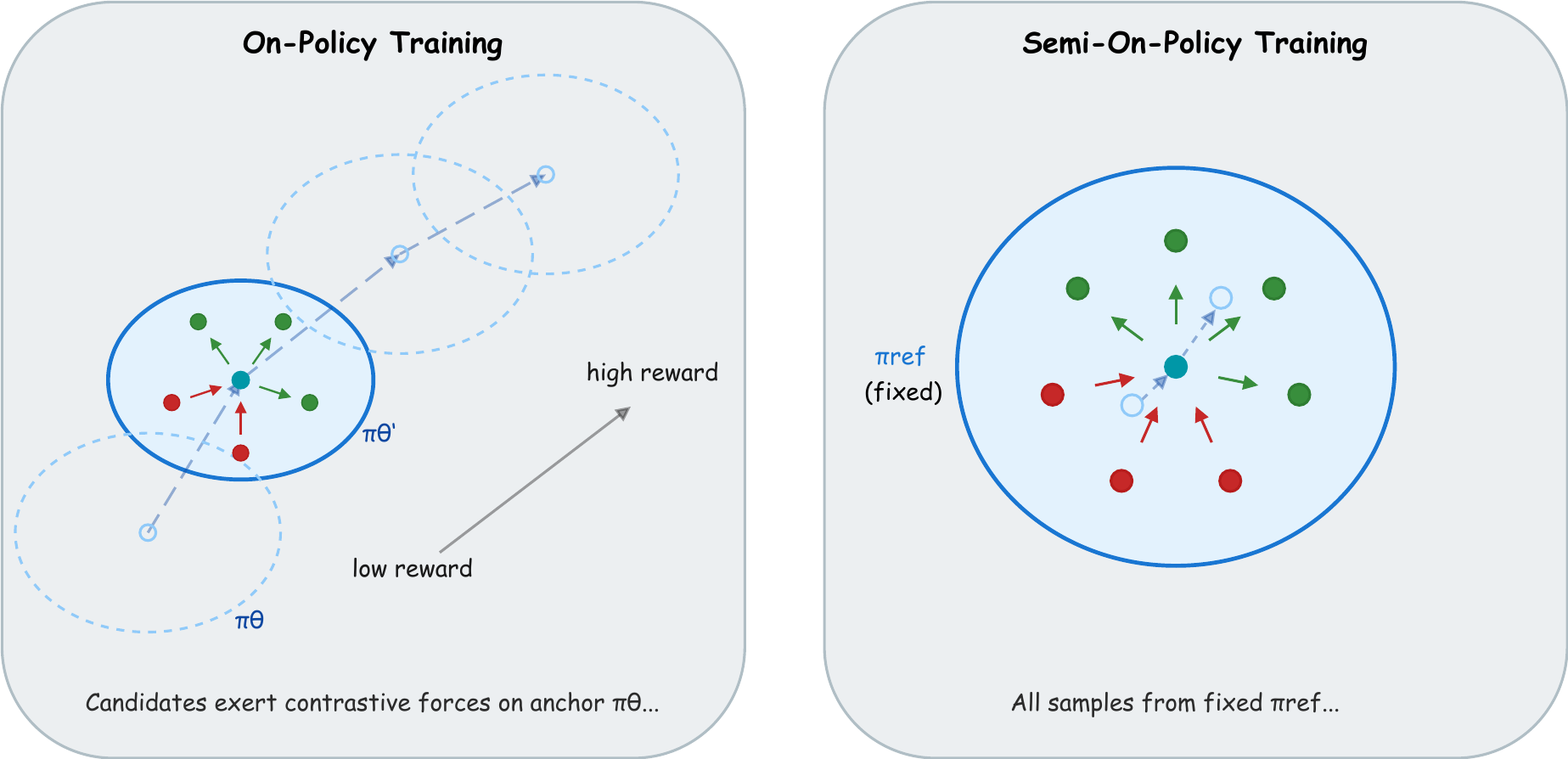}
    \caption{On-policy \vs semi-on-policy training under AR-CoPO. \textbf{Left:} On-policy training rolls out fresh candidates from the evolving policy $\pi_\theta$ at each iteration, enabling active exploration of new generation modes guided by the reward signal. \textbf{Right:} Semi-on-policy training fixes all rollouts to a reference policy $\pi_{\mathrm{ref}}$; the contrastive objective upweights high-reward candidates and suppresses low-reward ones within a trust region maintained by ratio clipping, enhancing exploitation without sacrificing stability. Each paradigm trains an independent LoRA adapter; merging the two adapters yields the final aligned model that benefits from both exploration and exploitation.}
    \label{fig:semi-on-policy}
  \end{figure}
  
  \subsection{Semi-On-Policy Alignment}

  \textbf{Limitations of pure on-policy exploration.}
  The on-policy AR-CoPO framework described above generates diverse candidate rollouts via initial noise perturbation, enabling active exploration of the generation space.
  However, not all reward signals respond equally to exploration-driven training.
  In particular, text alignment (TA)—a global, semantic-level reward that measures whether the generated video faithfully reflects the input prompt—is inherently difficult to improve through local noise perturbation alone.
  Because TA captures high-level semantic coherence that emerges from the video as a whole, small perturbations to the initial chunk noise typically yield semantically similar outputs with negligible reward variance, providing a weak and noisy gradient signal.
  This observation is consistent with prior work on aligning bidirectional video generators~\cite{xue2025dancegrpo, li2025branchgrpo}, where training toward TA reward is often omitted altogether due to instability or reward hacking.
  We argue that the root cause is a fundamental mismatch: exploration-based gradients operate locally in noise space, while semantic quality is a global property of the full sequence that cannot be reliably shaped by random perturbations.

  \textbf{Semi-on-policy training as exploitation.}
  To address this, we complement on-policy exploration with a dedicated \emph{exploitation} paradigm.
  As illustrated in Figure~\ref{fig:semi-on-policy} (right), rather than rolling out from an evolving policy, we fix all rollouts to a reference policy $\pi_{\mathrm{ref}}$ (the initialization checkpoint) and pre-collect a large replay buffer of reference candidates.
  This reference model already captures a reasonable generation distribution.
  The contrastive AR-CoPO objective is then applied over these fixed rollouts: high-reward candidates are upweighted while low-reward candidates are suppressed, \emph{without relying on stochastic exploration to discover new generation modes}.
  Concretely, the replay buffer stores chunk-level entries of the form $(h_{p-1},\,\epsilon_p^*,\,\{\epsilon_p^{(i)}\}_{i=1}^G,\,\{x_p^{(i)}\}_{i=1}^G,\,\{r^{(i)}\}_{i=1}^G)$.
  At each gradient step, we retrieve a batch of these entries, recompute the anchor prediction $\hat{x}_{0,t}^{(\theta)}$ from the current policy, and optimize the AR-CoPO objective in Eq.~\ref{eq:neighbor_obj} using the cached candidates and rewards.

  \textbf{Trust region via ratio clipping.}
  A na\"{i}ve off-policy scheme—applying the contrastive objective to fixed reference rollouts without any constraint—risks distributional shift: the policy may drift far from the reference distribution and collapse.
  We prevent this by retaining the ratio clipping in the AR-CoPO objective, which enforces a trust region around $\pi_{\mathrm{ref}}$~\cite{schulman2015trpo}.
  This constraint ensures that the policy's response to any single reference rollout remains bounded, maintaining training stability even when the buffer data becomes stale.
  We empirically confirm this in Section~\ref{sec:exp:semi}: removing ratio clipping causes rapid performance degradation, while the clipped variant preserves generation quality while improving VBench scores.

  \textbf{Combining exploration and exploitation via LoRA merging.}
  The on-policy and semi-on-policy objectives serve complementary purposes and are best optimized independently.
  We therefore train two separate LoRA adapters—one under on-policy AR-CoPO (for exploration and reward improvement) and one under semi-on-policy AR-CoPO (for exploitation and overall quality)—and merge them at inference time.
  As shown in Figure~\ref{fig:semi-on-policy}, this separation avoids interference between the two objectives during training.
  The merged model benefits from both paradigms: the semi-on-policy adapter reshapes the in-distribution quality of the reference policy, while the on-policy adapter steers the model toward higher reward regions through active exploration.

  \section{Experiments}
  
  We evaluate AR-CoPO on Self-Forcing~\cite{huang2025selfforcing}, which is the baseline model of many variants. We also evaluate Causal-Forcing~\cite{zhu2026causalforcing} and report the results in the supplementary material. We further report representative few-step streaming AR video generators as strong baselines such as LongLive~\cite{yang2025longlive}.
  Training is conducted on MovieGen Video Bench~\cite{polyak2024movie}.
  We optimize toward the VideoAlign~\cite{liu2025videoalign} reward suite, which comprises text alignment (TA), video quality (VQ), and motion quality (MQ).
  For the main experiments, all three rewards (TA, MQ, VQ) are jointly optimized.
  For holistic evaluation, we additionally report VBench scores (Quality, Semantic, Total).
  All models are fine-tuned with LoRA (rank 64, $\alpha=128$) using 24 GPUs.
We use a group size of $G=12$ and a learning rate of $1\times10^{-5}$.
Following Neighbor GRPO~\cite{he2025neighborgrpo}, we adopt symmetric anchor sampling with an anchor batch size of 4.
For the semi-on-policy strategy, we collect a replay buffer of 100 rollout groups from the initialization model.

  \begin{table}[t]
    \centering
    \caption{Quantitative comparison of streaming AR video generation models on VBench (Quality, Semantic, Total) and VideoAlign (VQ, MQ, TA, Overall) benchmarks. ``+ ours (Semi)'' denotes the model after semi-on-policy alignment; ``+ ours (Merged)'' denotes the final model obtained by merging the semi-on-policy and on-policy LoRA adapters.}
    \resizebox{0.7\linewidth}{!}{%
    \begin{tabular*}{0.9\linewidth}{@{\extracolsep{\fill}}lrrrrrrr}
    \toprule
    & \multicolumn{3}{c}{\textbf{VBench}} & \multicolumn{4}{c}{\textbf{VideoAlign}} \\
    \cmidrule(lr){2-4} \cmidrule(lr){5-8}
    \textbf{Method} & \textbf{Quality} & \textbf{Semantic} & \textbf{Total} & \textbf{VQ} & \textbf{MQ} & \textbf{TA} & \textbf{Overall}\\
    \midrule
    Self-Forcing     & 84.87 & 71.27 & 82.15 & 3.80 & 1.68 & 2.28  & 7.76\\
    Causal-Forcing & 85.27 & 70.35 & 82.28 & 3.97 & 1.43 & 2.40 & 7.79 \\ 
    LongLive & 85.10 & 71.16 & 82.31 & 3.87 & 1.76 & 2.43  & 8.06 \\
    \midrule
    Self-Forcing     & 84.87 & 71.27 & 82.15 & 3.80 & 1.68 & 2.28  & 7.76\\
    + ours (semi) & 85.15 & 71.68 & 82.45 &3.70 & 1.60 & 2.30 & 7.61 \\
    + ours (on-policy) & 84.81 & 70.71 & 81.99 & 4.15 & 2.06 & 2.30 & 8.51 \\
    + ours (merged) & 85.07 & 70.55 & 82.17 & 4.00 & 1.86 & 2.36 &  8.22 \\
    \bottomrule
    \end{tabular*}}
    \label{tab:comp:ar-models}
  \end{table}
  
  \subsection{Comparison with AR Models}
  
  \subsubsection{Quantitative results.}
  Table~\ref{tab:comp:ar-models} reports the main quantitative comparison.
  Semi-on-policy training alone surpasses all streaming AR baselines on VBench Total (82.45 vs.\ 82.31 for LongLive), with consistent gains on both Quality and Semantic dimensions, demonstrating the effectiveness of exploitation-focused alignment.
  After merging the on-policy LoRA adapter, VideoAlign Overall improves from 7.76 to 8.22.
  Crucially, this improvement is accompanied by a maintained VBench Total (82.15$\to$82.17), confirming that the gain reflects genuine alignment rather than in-domain score inflation.
  We elaborate on why this dual-benchmark criterion is the appropriate measure of alignment quality in Section~\ref{sec:exp:lora}.

  \begin{figure}[t]
    \centering
    \begin{subfigure}{0.9\linewidth}
      \includegraphics[width=0.99\textwidth]{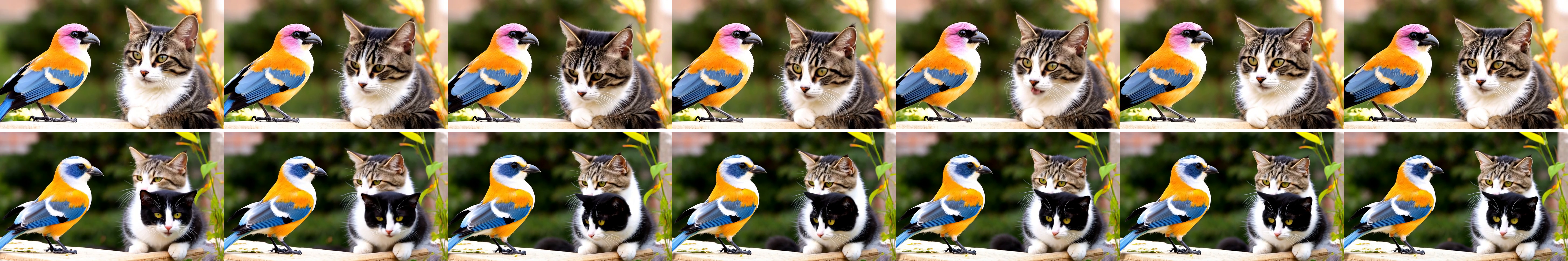}
        \caption{Prompt: A bird and a cat.}
    \end{subfigure}
    \\
    \begin{subfigure}{0.9\linewidth}
      \includegraphics[width=0.99\textwidth]{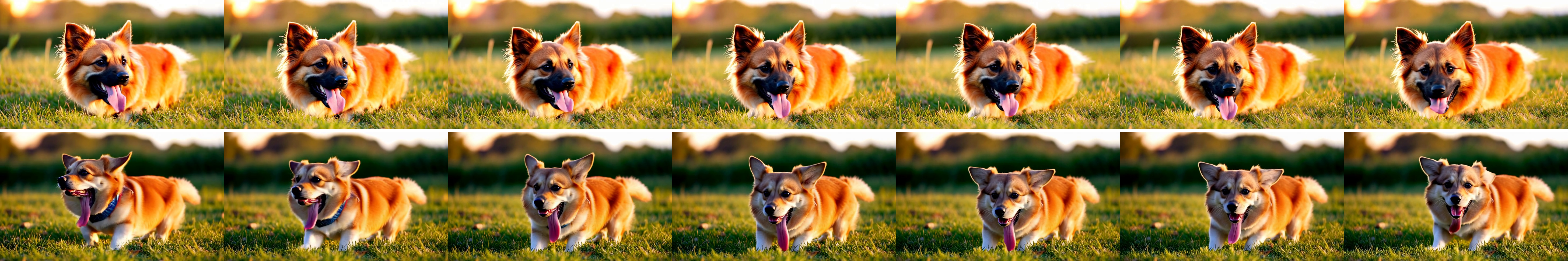}
        \caption{Prompt: A cute happy Corgi playing in park, sunset.}
    \end{subfigure}
    \\
    \begin{subfigure}{0.9\linewidth}
      \includegraphics[width=0.99\textwidth]{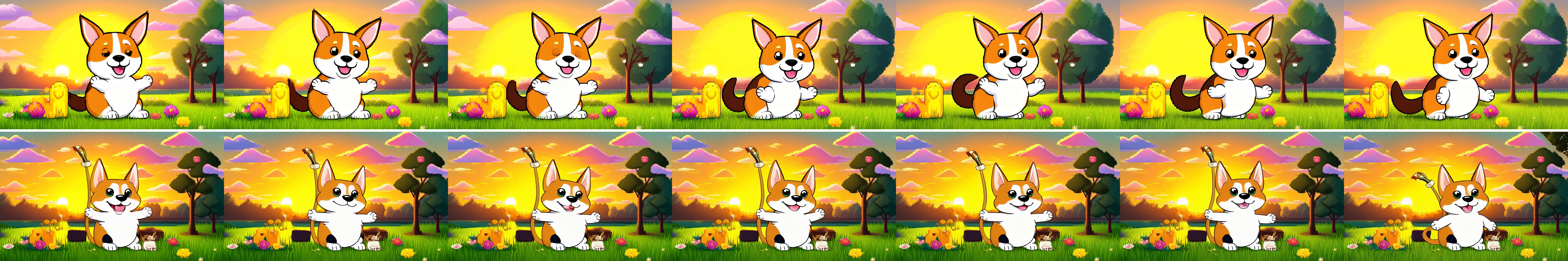}
        \caption{Prompt: A cute happy Corgi playing in park, sunset, pixel art.}
    \end{subfigure}
    \caption{Qualitative comparison between AR-CoPO (up) and Self-Forcing (down) on diverse text prompts. AR-CoPO produces videos with improved visual fidelity, motion quality, and better adherence to the text prompt.}
    \label{fig:cmp-quality}
  \end{figure}

  \subsubsection{Qualitative results.}
  Figure~\ref{fig:cmp-quality} shows side-by-side frame comparisons between AR-CoPO and Self-Forcing across diverse prompts.
  AR-CoPO produces videos with better aesthetic quality, more vivid appearance, more coherent motion, and better adherence to the text description.

\begin{figure*}[t]
    \centering
    \begin{subfigure}{0.79\linewidth}
      \includegraphics[width=0.99\textwidth]{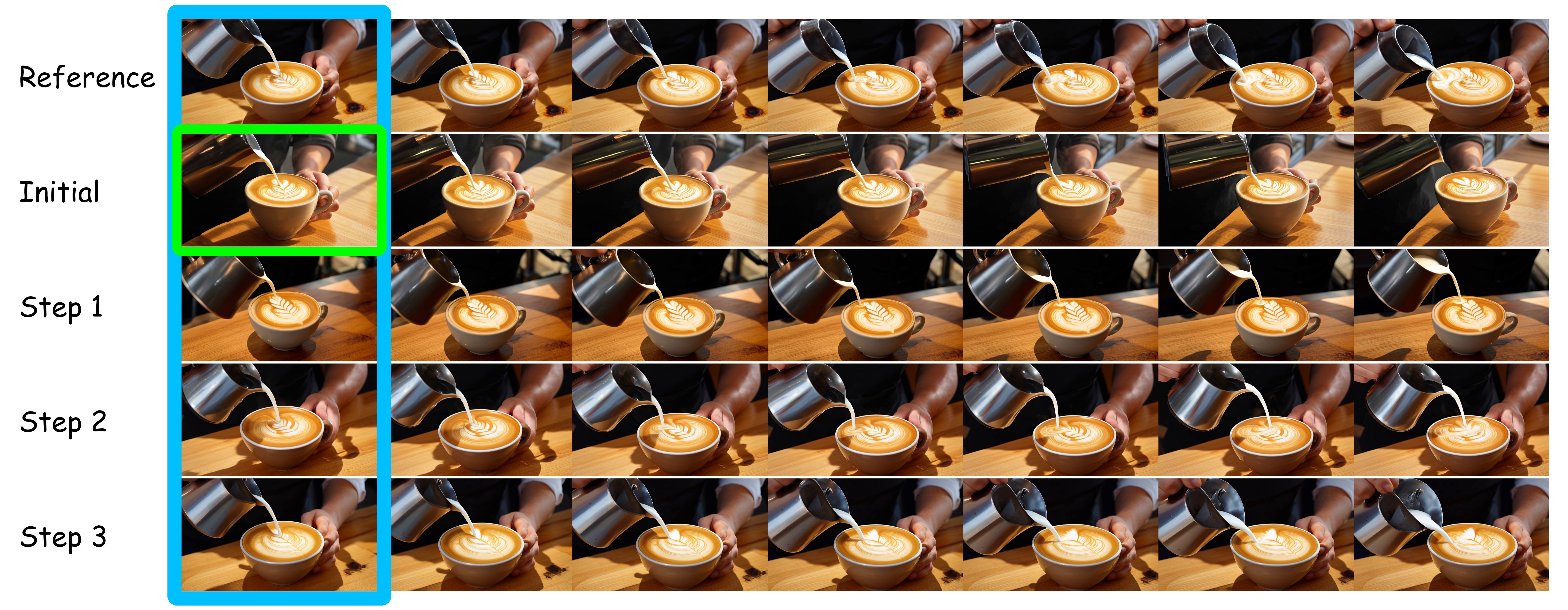}
        \caption{Fork@Chunk-1.}
        \label{fig:fork-randomness-1}
    \end{subfigure}
    \begin{subfigure}{0.79\linewidth}
      \includegraphics[width=0.99\textwidth]{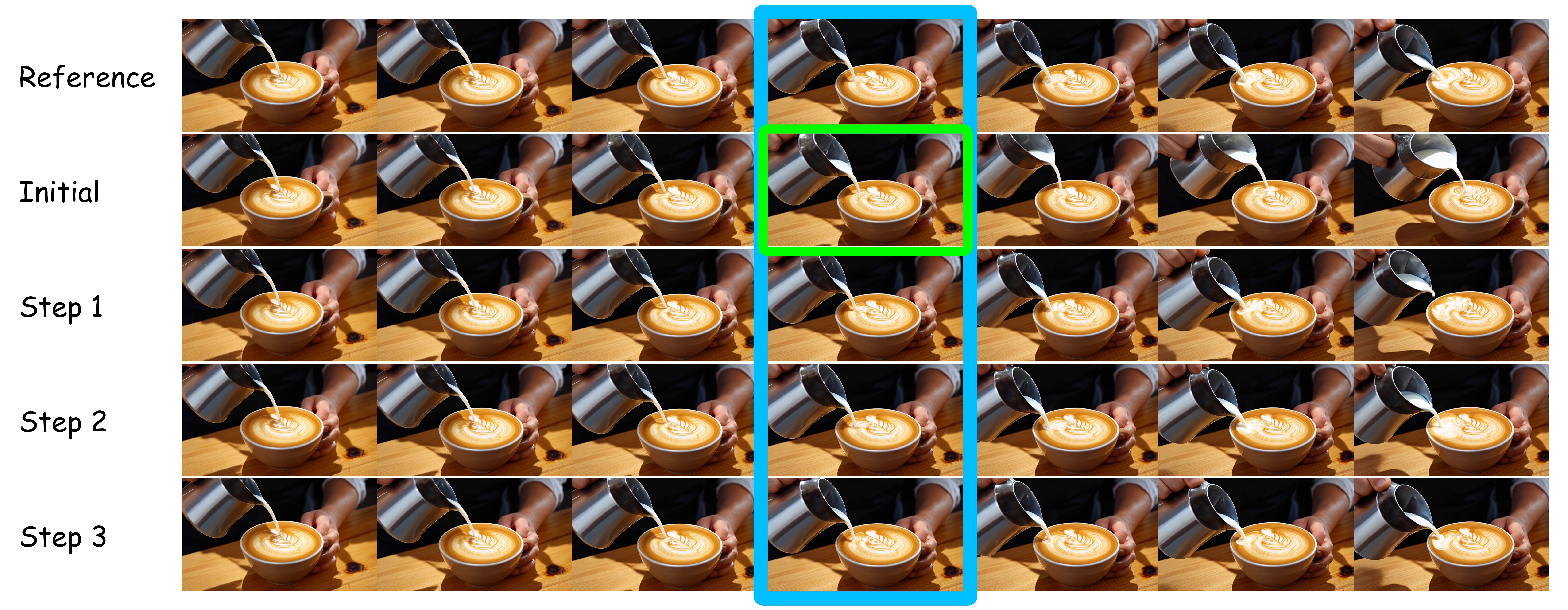}
        \caption{Fork@Chunk-4.}
        \label{fig:fork-randomness-4}
    \end{subfigure}
    \caption{Analysis of entropy sources in Self-Forcing. Each sub-figure corresponds to forking at a different chunk position. \textbf{Row 1:} Reference sample with all noise frozen. \textbf{Row 2:} Only the initial noise of the forked chunk is replaced—the output changes substantially. \textbf{Rows 3–5:} Only the CM solver noise at a specific denoising timestep within the chunk is replaced—the output changes marginally. This confirms that sample diversity in Self-Forcing is governed almost entirely by the initial noise, making intermediate SDE-style noise injection ineffective as an exploration mechanism.}
    \label{fig:fork-randomness}
  \end{figure*}


\begin{wrapfigure}{r}{0.4\linewidth}
  \centering
  \vspace{-1cm}
  \includegraphics[width=0.9\linewidth]{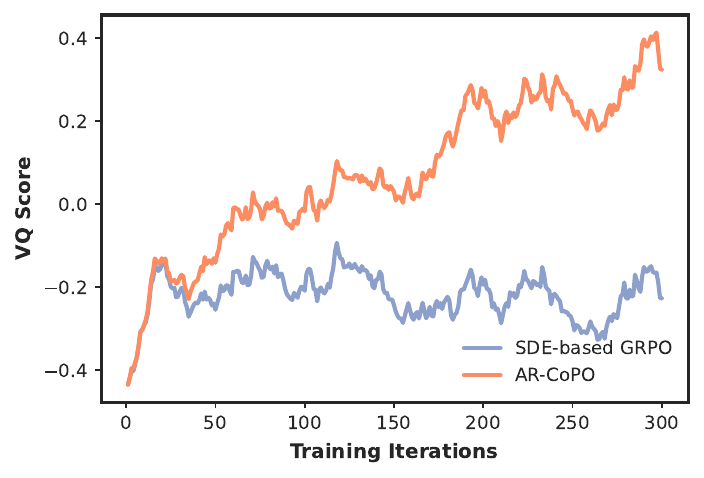}
  \caption{VQ training curves for AR-CoPO (ours) and the SDE-based GRPO baseline.\vspace{-0.5cm}}
  \label{fig:sde-vs-arcopo:vq}
\end{wrapfigure}

  \subsection{Comparison with SDE-GRPO}

  We compare AR-CoPO against an SDE-based GRPO baseline following the design of DanceGRPO~\cite{xue2025dancegrpo} and FlowGRPO~\cite{liu2025flowgrpo}. Training curves are shown in Figure~\ref{fig:sde-vs-arcopo} and Figure~\ref{fig:sde-vs-arcopo:vq}.
  The SDE-based variant fails to improve the reward throughout training, whereas AR-CoPO steadily achieves higher scores.

  We attribute this failure to the near-deterministic nature of distilled few-step consistency models (CMs).
  Although CM solvers inject intermediate re-noise during inference, sample diversity in Self-Forcing~\cite{huang2025selfforcing} is almost entirely governed by the \emph{initial} chunk noise rather than the intermediate solver noise.
  In other words, the model behaves as an approximately deterministic mapping from initial noise to clean frames, with intermediate stochasticity playing a negligible role.

  To validate this, we conduct a controlled noise substitution study (Figure~\ref{fig:fork-randomness}): we freeze all random draws in the sampler and selectively replace individual noise tensors.
  Replacing only the initial chunk noise (Row~2) causes substantial variation in the output, while replacing the intermediate CM solver noise at any individual denoising step (Rows~3–5) produces almost no visible change.
  Crucially, this effect is more pronounced in later chunks, where the accumulated autoregressive context further constrains the generation and makes it even more sensitive to the initial noise.

  SDE-based GRPO methods~\cite{xue2025dancegrpo, liu2025flowgrpo, li2025mixgrpo, li2025branchgrpo} such as DanceGRPO typically freeze the initial noise and define the action space over intermediate SDE noise injections.
  Because these intermediate injections carry negligible entropy in few-step AR models, the resulting policy gradient signal is near-zero, explaining why SDE-based training cannot drive reward improvement in this setting.

  \begin{table}[t]
    \centering
    \caption{Ablation of training strategies on VideoAlign and VBench metrics.}
    \resizebox{0.7\linewidth}{!}{%
    \begin{tabular*}{0.9\linewidth}{@{\extracolsep{\fill}}lrrrrrrr}
    \toprule
    & \multicolumn{3}{c}{\textbf{VBench}} & \multicolumn{4}{c}{\textbf{VideoAlign}} \\
    \cmidrule(lr){2-4} \cmidrule(lr){5-8}
    \textbf{Method} & \textbf{Quality} & \textbf{Semantic} & \textbf{Total} & \textbf{VQ} & \textbf{MQ} & \textbf{TA} & \textbf{Overall} \\
    \midrule
    Self-Forcing     & 84.87 & 71.27 & 82.15 & 3.80 & 1.68 & 2.28  & 7.76\\
    \midrule
    on-policy      & 81.66 & 69.68 & 79.26 & 3.53 & 0.25 & 2.63 & 6.42  \\
    off-policy     & 69.78 & 60.84 & 67.99 & 2.22 & -0.15 & 2.16 & 4.23 \\
    semi-on-policy & 85.15 & 71.68 & 82.45 & 3.70 & 1.60 & 2.30 & 7.61 \\
    \bottomrule
    \end{tabular*}}
    \label{tab:ablation}
  \end{table}
  
  \subsection{Effect of Semi-On-Policy Training}
  \label{sec:exp:semi}

  To isolate the effect of each training paradigm on semantic alignment, we ablate three training strategies under the AR-CoPO objective optimizing \emph{only} the TA reward: on-policy, semi-on-policy, and fully off-policy (without ratio clipping).
  This controlled setting allows us to directly assess whether semi-on-policy training can improve TA without the confounding effects of simultaneously optimizing VQ and MQ.
  Results are summarized in Table~\ref{tab:ablation}.

\begin{figure}[t]
    \centering
    \begin{subfigure}{0.8\linewidth}
        \includegraphics[width=\linewidth]{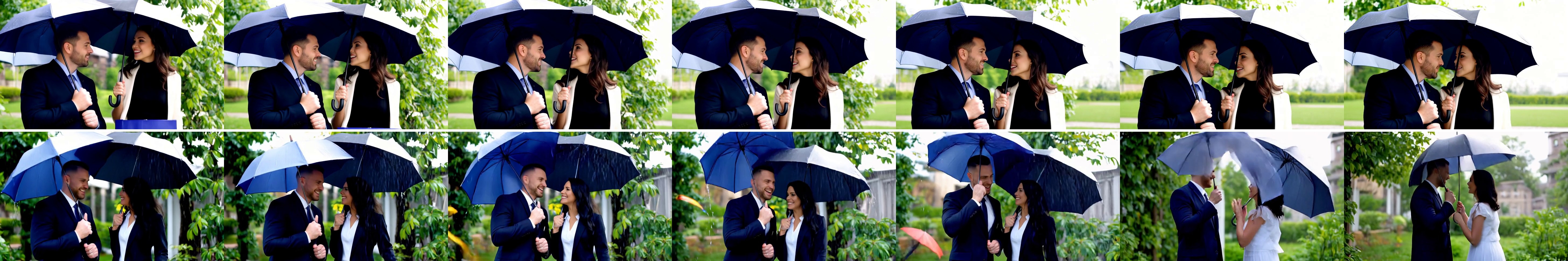}
        \caption{A couple in formal evening wear going home get caught in a heavy downpour with umbrellas.}
    \end{subfigure}
    \begin{subfigure}{0.8\linewidth}
        \includegraphics[width=\linewidth]{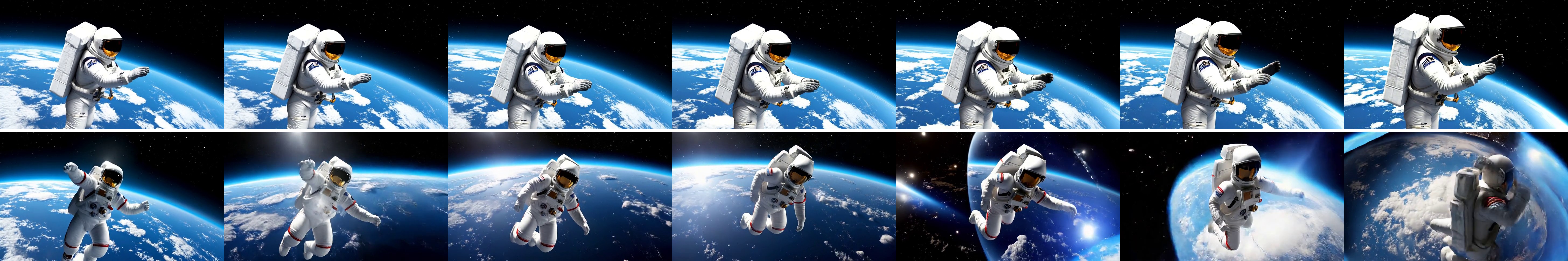}
        \caption{An astronaut flying in space, featuring a steady and smooth perspective.}
    \end{subfigure}
    \caption{Comparison of generated samples under semi-on-policy (top for each) and on-policy (bottom for each) training. On-policy training introduces visible temporal inconsistencies, whereas semi-on-policy training preserves generation quality.}
    \label{fig:ta-vs-ours}
\end{figure}

  \textbf{On-policy training} improves the in-domain TA score, but causes severe degradation across all other metrics and introduces visible generation artifacts (see Figure~\ref{fig:ta-vs-ours}).
  Most strikingly, MQ collapses from 1.68 to 0.25, accompanied by a sharp drop in VBench Total (82.15$\to$79.26).
  We identify this as a core failure mode of on-policy TA optimization.
  TA is a global, semantic-level reward: it measures whether the generated video as a whole faithfully reflects the input prompt.
  Optimizing such a signal via local noise-space exploration drives the model toward semantic shortcuts, \ie reward hacking, at the expense of temporal coherence and structural continuity.
  This manifests as the severe motion discontinuities and frame inconsistencies shown in Figure~\ref{fig:ta-vs-ours}, which directly cause MQ and VQ to collapse.
  This instability is consistent with prior work~\cite{xue2025dancegrpo, li2025branchgrpo}, where TA optimization is often omitted altogether due to similar failure modes.

  \textbf{Semi-on-policy training} avoids this collapse: VideoAlign scores remain broadly on par with the Self-Forcing baseline, while VBench Quality and Semantic scores improve over it.
  The ratio-clipping constraint is essential here---it keeps the policy within a trust region of the reference rollouts and prevents out-of-distribution drift.
  We confirm this by comparing against \textbf{fully off-policy training} (no ratio clipping): without the trust-region constraint, the model drifts far from the reference distribution, causing most scores to deteriorate.

  \subsection{LoRA Merging}

  \begin{table}[t]
    \centering
    \caption{Scores when merging on-policy weights with different strength}
    \resizebox{0.7\linewidth}{!}{%
    \begin{tabular*}{0.9\linewidth}{@{\extracolsep{\fill}}lrrrrrrr}
    \toprule
    & \multicolumn{3}{c}{\textbf{VBench}} & \multicolumn{4}{c}{\textbf{VideoAlign}} \\
    \cmidrule(lr){2-4} \cmidrule(lr){5-8}
    \textbf{Scale} & \textbf{Quality} & \textbf{Semantic} & \textbf{Total} & \textbf{VQ} & \textbf{MQ} & \textbf{TA} & \textbf{Overall} \\
    \midrule
    1.0    & 84.90 &  70.38  & 81.99 & 4.13 & 1.86 & 2.34 & 8.33 \\
    0.8    & 85.07 & 70.55 & 82.17 & 4.00 & 1.86 & 2.36 & 8.22  \\
    0.6    & 85.11 & 70.72 & 82.23 & 3.86 & 1.78 & 2.36  & 7.99\\
    0.4    & 85.14 & 71.44 & 82.40 & 3.76 & 1.62 & 2.34 &  7.72\\
    \midrule
    0 (Semi) & 85.15 & 71.68 & 82.45 & 3.70 & 1.60 & 2.30 & 7.61 \\
    \bottomrule
    \end{tabular*}}
    \label{tab:lora-merging}
  \end{table}

  The on-policy and semi-on-policy adapters are trained independently and merged at inference time by scaling the on-policy LoRA weights before addition.
  As shown in Table~\ref{tab:lora-merging}, varying the scale reveals a clear monotonic trade-off: increasing the on-policy contribution monotonically improves VideoAlign Overall (7.61$\to$8.33) while simultaneously degrading VBench Total (82.45$\to$81.99), and vice versa.
  This reflects the complementary nature of the two adapters---the semi-on-policy adapter exploits high-quality reference rollouts to preserve generation quality, whereas the on-policy adapter drives reward-seeking exploration that improves human preference scores.

  \textbf{Scale selection criterion.}
  \label{sec:exp:lora}
  As discussed in Section~\ref{sec:exp:semi}, directly optimizing AR video generators toward reward models is prone to in-domain reward hacking: the on-policy TA experiment demonstrates that a higher reward score does not always indicate better generation quality.
  Consequently, a higher VideoAlign Overall score alone is an unreliable criterion for model selection.
  We instead adopt a stricter criterion: the selected configuration must improve \emph{both} the in-domain benchmark (VideoAlign Overall) and the independent out-of-domain benchmark (VBench Total) relative to the baseline.
  Among all scales in Table~\ref{tab:lora-merging}, scale$=1.0$ achieves the highest VideoAlign Overall (8.33) but at the cost of VBench Total degradation (81.99 vs.\ 82.15 for the baseline), which we interpret as a sign of over-optimization rather than genuine quality improvement.
  Scale$=0.8$ is the largest scale that satisfies the dual-improvement criterion---VideoAlign Overall rises from 7.76 to 8.22 while VBench Total is maintained (82.15$\to$82.17)---and is therefore selected as the default reported setting.

  \section{Conclusion}
  In this paper, we present AR-CoPO, a chunk-level contrastive policy optimization framework for aligning few-step streaming autoregressive video generators to human preference.
  By constructing neighborhood candidates through a forking mechanism at a randomly selected pivot chunk, AR-CoPO circumvents the fundamental mismatch between SDE-based exploration and the near-deterministic dynamics of consistency model samplers.
  The complementary semi-on-policy strategy further improves generation quality by exploiting high-quality reference rollouts within a trust region, without sacrificing the exploration benefits of on-policy training.
  Experiments on Self-Forcing demonstrate consistent improvements on both VBench and VideoAlign, validating the effectiveness of AR-CoPO for post-training alignment of streaming AR video generation.

\section*{Acknowledgements}

This work is supported in part by NSFC-RGC Project N\_CUHK498/24, \\in part by Guangdong Basic and Applied Basic Research Foundation \\(No.~2023B1515130008, XW), in part by Shenzhen Loop Area Institute, and in part by the Centre for Perceptual and Interactive Intelligence (CPII) Ltd under the Innovation and Technology Commission (ITC)'s InnoHK.

%
%
\bibliographystyle{splncs04}
\bibliography{main}
\end{document}